# Hierarchical Partitioning of the Output Space in Multi-label Data

Yannis Papanikolaou, Ioannis Katakis, Grigorios Tsoumakas, *Affiliate, IEEE CS*

**Abstract**—Hierarchy Of Multi-label classifiERs (HOMER) is a multi-label learning algorithm that breaks the initial learning task to several, easier sub-tasks by first constructing a hierarchy of labels from a given label set and secondly employing a given base multi-label classifier (MLC) to the resulting sub-problems. The primary goal is to effectively address class imbalance and scalability issues that often arise in real-world multi-label classification problems. In this work, we present the general setup for a HOMER model and a simple extension of the algorithm that is suited for MLCs that output rankings. Furthermore, we provide a detailed analysis of the properties of the algorithm, both from an aspect of effectiveness and computational complexity. A secondary contribution involves the presentation of a balanced variant of the $k$ means algorithm, which serves in the first step of the label hierarchy construction. We conduct extensive experiments on six real-world datasets, studying empirically HOMER's parameters and providing examples of instantiations of the algorithm with different clustering approaches and MLCs, The empirical results demonstrate a significant improvement over the given base MLC.

**Index Terms**—Knowledge discovery, Machine learning, Supervised learning, Text mining

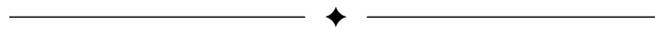

## 1 INTRODUCTION

IN multi-label learning, training examples are associated with a vector of binary target variables, also known as labels. The goal is to construct models that, given a new instance, predict the values of the target variables (classification), order the target variables from the most to the least relevant one with the given instance (ranking), do both classification and ranking, or even output a joined probability distribution for all target variables.

In the past decade multi-label learning has attracted a great deal of scientific interest. One main reason behind this is that a number of real-world applications can be formulated as multi-label learning problems; functional genomics [1], recommending bid phrases to advertisers [2], image [3] and music [4] classification are some example domains. The other main reason relates to the interesting challenges it poses, such as the identification and exploitation of dependencies among the target variables, the power-law distribution that the frequency of labels exhibits in several real-world applications and the increased space and time complexity involved in learning from multi-label data, especially when the number of labels is large.

This article presents a multi-label learning algorithm that we call HOMER[1] (Hierarchy Of Multi-label learnERs). HOMER is a divide-and-conquer algorithm, as it recursively partitions the vector of target variables into smaller disjoint vectors forming a hierarchy of such vectors. We employ a novel approach to perform this partitioning by clustering the labels using as a similarity measure the training examples for which they co-occur (more specifically we represent each label as a binary vector of its occurrences in the training set). This partitioning results in simpler learning tasks with fewer training examples (and features in the case of documents) and less evident class imbalance.

HOMER was first presented in [5], a technical report that was accepted for presentation at the Mining Multidimensional Data workshop of ECML PKDD 2008 in Berlin[2]. Since then, HOMER has been mentioned in several scientific papers[3]. It has been employed in diverse ways, such as for the automatic classification of edit categories in Wikipedia revisions [6], as a component of automated negotiation agents [7], for multi-label classification of economic articles [8] and for semantic-based recommender systems [9].

HOMER is a multi-label learning algorithm that achieves state-of-the-art prediction accuracy. An extensive experimental comparison involving 12 methods, 11 datasets and 16 evaluation measures concluded that HOMER is among the two best performing methods overall [10]. Another empirical comparison involving 8 methods, 11 datasets and focusing on the empty prediction rate, found HOMER among the two best performing methods too [11].

The contributions of this article that are inherited from the original technical report are:

- A novel multi-label classification algorithm that automatically constructs a hierarchy of sets of labels, learns a local multi-label classification model at every node of the hierarchy, and applies these models

---

- *Yannis Papanikolaou and Grigorios Tsoumakas are with the School of Informatics, Aristotle University of Thessaloniki, Greece. E-mail: {ypapanik,greg}@csd.auth.gr*
- *Ioannis Katakis is with the Department of Informatics and Telecommunications, National and Kapodistrian University of Athens. E-mail: katak@di.uoa.gr*

*Manuscript received ...*

1. Homer was an ancient Greek epic poet, best known as the author of Iliad and Odyssey (https://en.wikipedia.org/wiki/Homer).

2. The web page of the workshop is no longer available, but the papers that got accepted are listed in the workshop's page within the conference's site (http://www.ecmlpkdd2008.org/workshop-papers-mmd).

3. At the time of writing, Google Scholar reports 197 citations (https://scholar.google.gr/scholar?cluster=16386130204802114854).

hierarchically, in a top-down manner, to deliver predictions for new instances (Section 4). HOMER leads to state-of-the-art accuracy results and reduced time complexity during prediction compared to the standard one-vs-rest (also known as binary relevance) approach.
- An extension of the $k$ means algorithm, called *balanced $k$ means*, which produces equally-sized partitions (Section 3). Balanced $k$ means is used recursively in the first step of HOMER in order to construct the hierarchy of labelsets, leading to better results compared to non-balanced clustering approaches.

Besides serving as an archival publication for HOMER, this article contributes the following novel and significant extensions to the original paper:

- The addition of a parameter for controlling the expansion of the label hierarchy, which generalizes the original description of HOMER, and allows it to perform better in domains with many rare labels (Section 4).
- A direct extension of the algorithm, to account for algorithms that output rankings as results, or for scenarios where the desired output is as well a ranking (Section 4.2).
- A detailed complexity analysis for the algorithm (Section 4.3).
- A short discussion on what are the aspects that should be taken into account to construct an effective HOMER model and an analysis of how HOMER performs with respect to rare and frequent labels (Section 4.4).
- Extensive empirical comparisons on six real world corpora, to analyze and study the algorithm's parameters behavior, propose different instantiations of HOMER's components and assess the improvement over the given base MLC (Section 5).
- As a secondary contribution, we provide a review of similar approaches and resolve a number of misconceptions around HOMER in the literature (Section 2).

## 2 RELATED WORK

The key idea in HOMER is the automatic construction of a hierarchy on top of the labels of a multi-label learning task. While this was novel at that time within the multi-label learning literature, the same idea had already been studied for the single-target multi-class classification task [12], [13]. In both of these approaches, the similarity between classes is based on their average feature vector (centroid). In HOMER, in contrast, each label is represented as a binary vector whose dimensions correspond to the training examples and whose values indicates whether the corresponding training example is annotated with the label. Calculating label similarity based on this vector space would not make sense in the multi-class case, but it does in the multi-label case, where labels are overlapping, and can co-occur at the same training example. In [12], similarity was measures on a set of discriminative features selected based on the Fisher index, while in [13], similarity was measured in a lower-dimensional feature space obtained through linear discriminant projection. As far as the hierarchy construction process is concerned, in [12], this was done top-down using spherical 2-means, initializing the algorithm with the two farthest classes. In [13], it was done bottom-up using agglomerative hierarchical clustering. In HOMER, in contrast, the use of balanced $k$ means is another key difference, which can lead to balanced trees (not necessarily binary) that offer guarantees with respect to prediction complexity.

Another view of HOMER is that it addresses a multi-label task by breaking down the entire label set (recursively) into several disjoint smaller ones. A similar pattern, but randomly and non-recursively, is followed in the disjoint version of Random $k$ Labelsets (RA$k$EL$_d$) [14]. RA$k$EL$_d$ was extended in [15], by introducing an algorithm that divides the label set into several mutually exclusive subsets by taking into account the dependencies among the labels, instead of randomly.

HOMER is a meta-algorithm, in the sense that it employs a base MLC on each of the sub tasks it creates out of the initial task. However, this perspective is sometimes overlooked in the literature [15], [16], [17], where HOMER is perceived just as its default instantiation using binary relevance as the multi-label learner with C4.5 trees as binary classifiers. Another misconception in the literature, is that it is erroneously considered as a label-powerset method [18], [19].

A variation of HOMER, where the calibrated label ranking algorithm was used as MLC was proposed in [20]. Finally, three different algorithms (balanced $k$ means, predictive clustering trees (PCTs) and hierarchical agglomerative clustering) for constructing the label hierarchy of HOMER were studied in [21] using the random forest of PCTs as the MLC.

## 3 BALANCED $k$ MEANS

Before proceeding with the presentation of HOMER, we describe an extension of the $k$ means clustering algorithm, called balanced $k$ means, which sets an explicit constraint on the size of each cluster. Let us denote as $S$ the set of the data points to be clustered and $\lambda$ a given data point with $W_\lambda$ the set of relevant data vectors, $k$ being the number of partitions and $it$ the number of iterations. $N_\lambda$ will denote the dimensionality of the data points. Algorithm 1 shows the relevant pseudo-code.

The key element in the algorithm is that for each cluster $i$ we maintain a list of data points, $C_i$, sorted in ascending order of distance to the cluster centroid $c_i$. When the insertion of a point into the appropriate position of the sorted list of a cluster, causes its size to exceed the maximum allowed number of points (approximately equal to the number of items divided by the number of clusters), the last (furthest) element in the list of this cluster is inserted to the list of the next most proximate cluster. This may lead to a cascade of $k - 1$ additional insertions in the worst case. As opposed to $k$ means, we limit the number of iterations using a user-specified parameter, $it$, as no investigation of convergence was attempted.

**ALGORITHM 1:** Balanced $k$ means Algorithm

**Input:** number of clusters $k$, data $S$, data vectors $W_\lambda$, iterations $it$
**Output:** $k$ balanced clusters of $S$
**for** $i \leftarrow 1$ **to** $k$ **do**
  // initialize clusters and cluster centers
  $C_i \leftarrow \emptyset$ ;
  $c_i \leftarrow$ random member of $S$ ;
**end**
**while** $it > 0$ **do**
  **foreach** $\lambda \in S$ **do**
    **for** $i \leftarrow 1$ **to** $k$ **do**
      $d_{\lambda i} \leftarrow$ distance$(\lambda, c_i, W_i)$
    **end**
    finished $\leftarrow$ false;
    $\nu \leftarrow \lambda$ ;
    **while** *not finished* **do**
      $j \leftarrow \arg\min_i d_{\nu i}$;
      Insert sort $(\nu, d_\nu)$ to sorted list $C_j$;
      **if** $|C_j| > \lceil |S|/k \rceil$ **then**
        $\nu \leftarrow$ remove last element of $C_j$;
        $d_{\nu j} \leftarrow \infty$ ;
      **end**
      **else**
        finished $\leftarrow$ true;
      **end**
    **end**
  **end**
  recalculate centers;
  $it \leftarrow it - 1$
**end**
**return** $C_1, ..., C_k$;

### 3.1 Computational Complexity

At each iteration of the balanced $k$ means algorithm, we loop over all points of $S$, calculate their distance to the $k$ cluster centers with an $\mathcal{O}(|N_\lambda|)$ complexity and insert them into a sorted list of max size $|S|/k$, which has complexity $\mathcal{O}(|S|)$. This may result into a cascade of $k-1$ additional insertions into sorted lists in the worst case, but the complexity remains $\mathcal{O}(|S|)$. So the total cost of the balanced $k$ means algorithm is $\mathcal{O}(|S||N_\lambda| + |S|^2)$. As typically $|S| \ll |N_\lambda|$, the algorithm can efficiently partition labels into balanced clusters based on very large datasets.

### 3.2 Related Work

A sampling-based algorithm with complexity $\mathcal{O}(|S|log|S|)$ has been proposed in [22]. The frequency-sensitive $k$ means algorithm [23] is a fast algorithm for balanced clustering (complexity of $\mathcal{O}(|S|)$). It extends $k$ means with a mechanism that penalizes the distance to clusters proportionally to their current size, leading to fairly balance clusters in practice. However, it does not guarantee that every cluster will have at least a pre-specified number of elements. Another approach to balanced clustering extends $k$ means by considering the cluster assignment process at each iteration as a minimum cost-flow problem [24]. Such an approach has a complexity of $O(|S|^3)$, which is worse than the proposed algorithm. Finally, according to [22], the balanced clustering problem can also be solved with efficient min-cut graph partitioning algorithms with the addition of soft balancing constraints. Such approaches have a complexity of $(O(|S|^2)$, similarly to the proposed algorithm.

## 4 HIERARCHY OF MULTI-LABEL CLASSIFIERS

In this section we describe HOMER, based on the initially presented algorithm in [5], along with a number of extensions to the previous work.

Before proceeding, we present the notation used throughout the paper. Let us define as $L$ the label set of the multi-label task that we wish to address and $l$ a label. Similarly, $D_{Train}$ and $D_{Test}$ will express the set of training and the set of test instances respectively and $d$ an instance. For simplicity, when referring to $D$, unless otherwise noted, we will mean $D_{Train}$. The set of non-zero features of $d$ will be defined as $f_d$ and the instance's labelset as $L_d$. As HOMER proceeds by constructing a hierarchy out of the dataset, we will represent the training set at each node as $D_n$ and the labels that are relevant to the node as $L_n$. Also, each node will have a set of meta-labels, $M_n$ (their role will be explained further on). Finally, we will refer to a multi-label learning classifier as $MLC$ and a clustering algorithm as $C$.

### 4.1 Description

A HOMER model is essentially a generic method to bundle any given multi-label classifier aiming to improve performance and computational complexity. The main idea is the transformation of a multi-label classification task with a large set of labels $L$ into a tree-shaped hierarchy of simpler multi-label classification tasks, each one dealing with a small number of labels. The algorithm consists of two parts, first the creation of a label hierarchy out of the label set and second the training and prediction locally at each node of the hierarchy, with a given $MLC$. Below, we describe these steps in detail.

#### 4.1.1 Label hierarchy

To construct a label hierarchy we first need to determine a vector representation for each label. A simple choice, is to represent each label $l$ as a binary vector $V_l$ of $|D_{Train}|$ dimensions, with

$$V_l(d) = \begin{cases} 1, & \text{if } l \in L_d \\ 0, & \text{otherwise} \end{cases} \quad (1)$$

The motivation is that labels that co-occur in instances will be more similar and thus more likely to belong to the same cluster. Upon selection of a proper distance function for the label vectors, we employ a clustering algorithm $C$ and perform an iterative clustering of labels, until each node has only a few labels (the initial HOMER algorithm in [5] partitions $L$ until each leaf node has only one label). Specifically, the procedure is as follows; starting from the root node of the hierarchy, and using the clustering algorithm, we partition the initial label set into a number of children-clusters. Each of the resulting clusters defines a new node, which is in turn partitioned into its children - clusters. A node's labels are not further partitioned if $|L_n| \leq nmax$, where $nmax$ is a user-defined threshold that specifies the maximum number of labels in the leaf nodes of the hierarchy. The initially

presented HOMER algorithm employed only $k$ means and balanced $k$ means as clustering algorithms, from the above description however, it becomes clear that it is possible to employ any given clustering algorithm for this task.

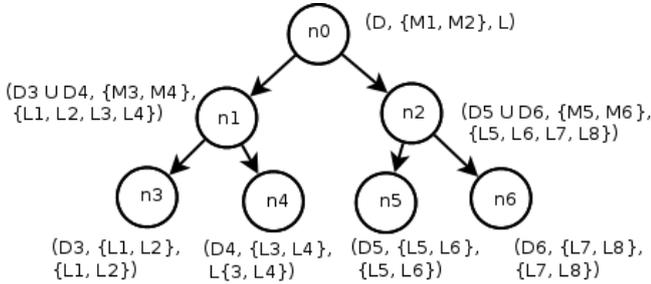

Fig. 1: Training for a simple HOMER model. At each node we depict in parentheses, the training set, the label set on which the $MLC$ is trained and the respective $L_n$

### 4.1.2 Training

For each node we train a local $MLC$ with $|D_n|$ instances. $D_n$ comprises all instances from $D$ annotated with at least one of the labels of $L_n$. The local classifier is trained on $M_n$ which will be either identical to $L_n$ (if the node is a leaf of the hierarchy) or a set of meta-labels, with each meta-label $\mu_c$ corresponding to one of the children nodes. Formally, we denote $M_n = \{\mu_c \mid c \in \text{children}(n)\}$ with $\mu_c$ having the following semantics: a training example can be considered annotated with $\mu_c$, if it is annotated with at least one of the labels of $L_c$. We would like to further clarify the difference between $L_n$ and $M_n$; the first set, is the set of the labels that the clustering algorithm assigned to the given node $n$ during the clustering process. The latter set, is the label set on which the MLC is trained. Figure 1 shows an example of a HOMER hierarchy, with the training set, $M_n$ and $L_n$ at each node. From the above description, it is easy to see that any given $MLC$ can be used for training each node.

### 4.1.3 Prediction

During prediction on new, unannotated data, each new instance $x$ is traversing the tree as follows: starting from the root node, the local MLC assigns to each instance zero, one or more meta-labels. Then, by following a recursive process $x$ is forwarded to those nodes that correspond to the assigned meta-labels. In other words, an instance $x$ is forwarded to a child node $c$ only if $\mu_c$ is among the predictions of the parent MLC. Eventually, $x$ reaches the leaves of the hierarchy. at which point the algorithm combines the predictions of the terminal nodes. Figure 2(a) illustrates the aforementioned prediction process for a simple HOMER model.

To summarize the description of HOMER, we provide the relevant pseudocode in Algorithm 2.

## 4.2 Extension for ranking MLCs

The original HOMER algorithm requires that the local $MLC$ outputs bipartitions and proceeds by combining the terminal nodes predictions into the overall predictions for a given new instance. Nevertheless, many multi-label algorithms

**ALGORITHM 2:** HOMER
**Input:** $D_{Train}, L, D_{Test}, C, MLC, nmax$

```
/* Clustering                              */
RecursiveClustering(L);
/* Training                                */
for each node n ∈ hierarchy do
    D_n = {d | d ∈ D, ∃l | l ∈ L_n ∧ l ∈ L_d};
    if n is a leaf node then
        M_n = L_n;
    end
    else
        M_n ={μ_c | c is a child of n};
    end
    train MLC_n on training set D_n, label set M_n;
end
/* Prediction                              */
for each d ∈ D_Test do
    RecursivePrediction(ROOT, d) ;
    /* ROOT is the root node of the hierarchy.
       */
    Predictions_d = ∪predictions_{LEAF-NODES};
end
/* Recursive Label clustering.             */
Procedure RecursiveClustering(Labels L_n)
    Cluster(L_n) into k children nodes with C;
    /* k does not need to be the same along
       iterations and is dependent of the C in
       use.                                  */
    for each child node n' do
        if |L_n'| > nmax then
            RecursiveClustering (L_n);
        end
    end
    return;

/* Recursively predict d, on the label
   hierarchy.                              */
Procedure RecursivePrediction(Node n, instance d)
    predict with MLC_n;
    if n ≠ leaf-node then
        for each μ_c ∈ M_n do
            if μ_c is assigned to d then
                RecursivePrediction(c, d);
            end
        end
    end
    return;
```

produce a ranking of labels for a new instance, requiring the choice of a thresholding technique to obtain a bipartition set. Furthermore, there exist scenarios where one needs to predict a ranking instead of a bipartitions set (for instance a search query). HOMER can be extended so as to account for the above cases with the following modification; during prediction, each node's MLC, instead of predicting one or more meta-labels for a given instance, assigns a score (or probability) to each of them, which is the ranking score (or probability) of the base MLC. Subsequently, the children nodes will propagate these scores, by multiplying their own predictions with the score they have been assigned. Figure 2 illustrates the process for both classic HOMER and the proposed extension. In order to avoid a full expansion of the tree, one can prune away a given node path, by applying some heuristics. In the experiments for instance, we employ this approach to train HOMER-LLDA models and prune

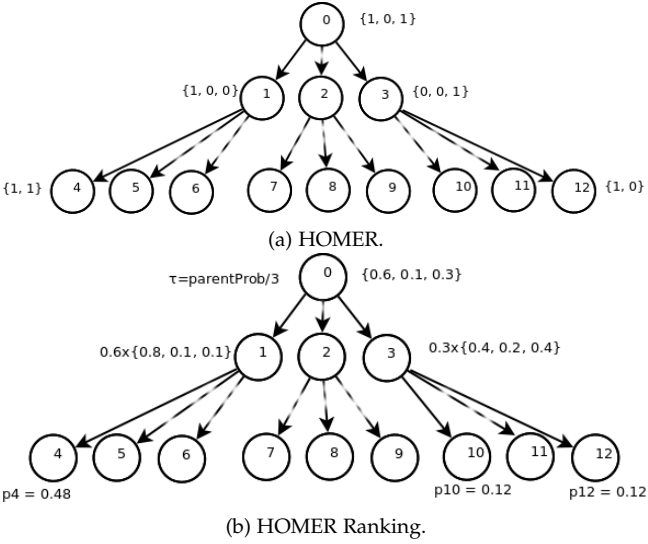

Fig. 2: Prediction process for (a) the classic HOMER algorithm and (b) the relevant extension to address ranking outputs. Specifically for (b) a pruning scheme has been followed, eliminating paths with a probability $p \leq \frac{parentProbability}{k}$.

TABLE 1: Complexities for the Employed Clustering Algorithms.

| Clustering Algorithm | Complexity |
| --- | --- |
| Balanced $k$ means | $\mathcal{O}(|L||D| + |L|^2)$ |
| Fast OPTICS | $\mathcal{O}(|D||L|\log^2|L|)$ |
| SLINK | $\mathcal{O}(|L|^2)$ |

away nodes having a probability $p \leq \frac{parentProbability}{k}$.

### 4.3 Computational Complexity

From the above description, HOMER's training complexity will be the combination of the clustering algorithm's complexity and the training cost of the hierarchy nodes:

$$f_H = f_{CLUST}(|L|,|D|) + f_{HTrain}(|L|,|D|) \quad (2)$$

The complexity of the balanced clustering process at each node $n$ depends on the actual algorithm being used and can range from $O(|L_n|)$ to $O(|L_n|^3)$ (see Section 3.2). $L_n$ is equal to $L$ at the root, but subsequently decreases exponentially with the tree's depth. Therefore, if $f(|L_n|)$ is a function of the complexity of the balanced clustering algorithm with respect to $L_n$, then the overall complexity of HOMER with respect to this algorithm is $O(f(|L|)$. In other words HOMER retains the complexity of the balanced clustering algorithm.

Consider for example that $f(|L_n|) = |L_n|^2$. Then at the root we have a cost of $|L|^2$ while at the second level we have $k$ additional costs of $(|L|/k)^2$, i.e. an additional cost of $|L|^2/k$. At the next level we have $k^2$ additional costs of $(|L|/k^2)^2$, i.e. an additional cost of $|L|^2/k^2$. This is a sum of a geometric series leading to a total cost of $2|L|^2$ when the depth of the tree approaches infinity.

In the experiments, we employ three different clustering algorithms, balanced $k$ means, FastOPTICS and SLINK.

Table 1 shows the relevant complexities for these three algorithms.

Concerning the second part of Equation 2, we will simplify our analysis by assuming that we employ a balanced clustering algorithm, with each node being partitioned to $k$ children. Let us denote the hierarchy depth with $\nu$. We further assume that $k^\nu = \frac{|L|}{nmax}$ and the complexity of the multi-label classifier that we employ is linear w.r.t $|D|$.

In this case, the hierarchy will have $\frac{\frac{|L|}{nmax}-1}{k-1}$ nodes (the sum of a geometric sequence). As described in the previous section, each of the terminal nodes of the hierarchy will have at most $nmax$ labels to train and predict, whereas any non-terminal node will have $k$ meta-labels respectively (the number of its children nodes). Therefore, we have

$$f_{HTrain} = \frac{k^{\nu-1}-1}{k-1} \times f(k, \overline{|D_{non-leaf}|}) + k^{\nu-1} \times f(nmax, \overline{|D_{leaf}|}) \quad (3)$$

or

$$f_{HTrain} = \frac{\frac{|L|}{nmax \times k}-1}{k-1} \times f(k) \times \overline{|D_{non-leaf}|} + \frac{|L|}{nmax \times k} \times f(nmax) \times \overline{|D_{leaf}|} \quad (4)$$

This is equivalent to

$$f_{HTrain} = (|L|\frac{1}{k(k-1) \times nmax} - \frac{1}{k-1}) \times f(k) \times \overline{|D_{non-leaf}|} + \frac{|L|}{nmax \times k} \times f(nmax) \times \overline{|D_{leaf}|} \quad (5)$$

Therefore, the training complexity of a HOMER model with balanced hierarchy will be

$$f_{HTrain} \in \mathcal{O}(|L|(\overline{|D_{non-leaf}|} + \overline{|D_{leaf}|})) \quad (6)$$

From the above we observe that HOMER's training complexity is linear with respect to $|L|$ regardless of the baseline classifier's complexity (here we have assumed that $k \ll |L|$ and $nmax \ll |L|$, a valid assumption for most real-world applications). With respect to $|D|$, HOMER also brings an improvement compared to the baseline algorithm's complexity. This improvement is difficult to be quantified though, as each node's training corpus is the union of it's labels occurrences in $D$ and thus it depends on a variety of factors, including the label frequencies, the overlap of labels in training instances and, most importantly, the quality of the label clustering.

Assuming again a balanced hierarchy, during prediction the complexity of the algorithm depends on the number of different paths that each new instance will take in the label hierarchy (for instance in Figure 2(a) the instance follows two different paths from a total of nine possible ones). Assuming that the MLC has a prediction complexity of $\mathcal{O}(f(|L|))$, then, in the ideal case where only one such path is followed, HOMER's prediction complexity will be

$$f_{HPrediction} = log_k(|L|) \times f(k) + f(nmax)$$

or

$$f_{HPrediction} \in \mathcal{O}(log_k(|L|)) \quad (7)$$

In the worst case, if all paths would be followed we would have

$$f_{HPrediction} = \frac{\frac{|L|}{nmax \times k} - 1}{k - 1} \times f(k) + \frac{|L|}{nmax \times k} \times f(nmax)$$

therefore

$$f_{HPrediction} \in \mathcal{O}(|L|) \quad (8)$$

4. We have used the assumption that $f_{MLC}$ is linearly dependent on $D$ to average over $D_n$.

## 4.4 Discussion

As explained in the introductory section of this paper, the motivation behind HOMER is to tackle a problem that would be difficult to solve globally, by breaking it to several local sub-problems which are expected to be more easily and effectively addressed. In this section, we focus on the factors that play an important role in building an effective HOMER model.

The most important part of the algorithm is the construction of a good label hierarchy. By 'good', we imply that the clusters should have as much as possible similar labels within them. A good hierarchy can engender the following benefits. First, similar labels will be expected to co-occur frequently. As a result, a cluster containing labels that are related, will tend to have a smaller training corpus than one containing dissimilar ones. As explained in the previous section, this leads in a shorter training time. A second benefit involves prediction; if the clusters of the hierarchy contain very similar labels, then, during prediction a new instance will follow only few (or ideally one) paths in the label hierarchy and therefore achieve a logarithmic complexity. A third benefit involves performance; a hierarchy with similar clusters will cause the MLC at each node to be more effective in predicting correctly the meta-labels for an unannotated instance.

Another substantial aspect in HOMER's configuration relates to the $nmax$ parameter. The initially presented HOMER model was expanding totally the label hierarchy tree, with terminal nodes having only a single label. In real world applications though, most labels tend to have very few positive examples and therefore very low frequencies. Full expansion of the tree in this case, would lead to very small training sets for each label and therefore poor performance. The model that we propose in this work, is using the $nmax$ parameter to address the above issues and stop the hierarchy expansion. As $nmax$ approaches $|L|$, the hierarchy will be shallower and the gain in performance smaller. On the other hand, as $nmax$ approaches to 1 the training sets of the nodes will be smaller and performance can even be worse than the baseline $MLC$. As a rule of thumb, we propose, depending on $|L|$ size, $|nmax|$ values in the order of 10 to $10^2$.

In order to better understand the role of $nmax$, and generally the differences among hierarchies with many or few nodes, let us inspect more closely how HOMER proceeds; as mentioned earlier, after creating a hierarchy of the label space, the algorithm trains each node with a subset of the training instances. More specifically, at each node the training set consists of the union of positive examples of each label. Hence, as we proceed to the nodes further down the hierarchy, we expect that this kind of sub-sampling will lead to learning problems that will have fewer and fewer negative examples for each label of the node (by definition, the number of positive examples will remain steady for each label). The same trend will be observed among HOMER models, as we increase the total number of nodes.

This is no random sub-sampling though; through the iterative clustering process (during the construction of the label hierarchy) we have put similar labels in the same cluster. Therefore, for each label, the negative examples will consist of the union of positive examples of the other similar labels, excluding of course the instances for which the labels co-occur. Through this process, we expect that the negative examples will provide a greater discriminative power to the MLC, in learning more accurately the task by distinguishing more effectively between similar labels.

This sub-sampling process will be expected to have a different effect on rare and frequent labels. Specifically, if we increase the total number of nodes in the hierarchy, rare labels will most probably benefit from reducing the imbalance between positive and negative examples. On the other hand, for frequent labels we expect at some point that there will be an 'inverse' imbalance with many positive examples and too few negative ones, in which case performance will likely drop.

In Section 5 we study empirically how the sub-sampling process influences performance of frequent and rare labels, validating our observations and remarks.

## 5 EXPERIMENTS

We performed five sets of experiments. The first experiment studies how frequent and rare labels are influenced by the total number of nodes in a HOMER model. In the second and third series of experiments, we investigated the role of parameters $k$ (number of cluster-children nodes) and $nmax$ (maximum number of labels in every leaf node) with respect to performance. In the above cases, we employed balanced $k$ means as the clustering algorithm and Binary Relevance with Linear SVMs as the baseline method.

In the fourth series of experiments we employed three different clustering algorithms using two different multi-label algorithms as base classifiers, Binary Relevance with Linear SVMs and Labeled LDA. Finally, we employed HOMER-BR on two large-scale corpora and compared the algorithm against the respective baseline (BR-SVM).

The code of the implementation and experiments is available at http://users.auth.gr/~ypapanik/.

### 5.1 Implementations and parameter setup

We used the ELKI library [25] for the clustering algorithms as well as for the Jaccard distance measure. The LibLinear package was employed for the Linear SVMs [26] in the Binary Relevance approach, keeping default parameters ($C = 1$, $e = 0.01$ and L1R-L2LOSS-SVC as a solver). Labeled LDA was implemented with the Prior-LDA variation [27]. For the latter, we used the Collapsed Gibbs Sampling method, with only one Markov Chain for simplicity and 100 iterations during training and prediction (with 50 iterations of burn-in period and a sampling interval of 5 iterations). Parameter $\beta$ was set to 0.1 while $\alpha$ was set to $\frac{50.0}{|L|}$ during training and $50.0 * \frac{frequency(l)}{sumOfFrequencies} + \frac{30.0}{|L|}$ during prediction (following the Prior-LDA approach). Specifically for Labeled LDA, as the Collapsed Gibbs Sampler follows a stochastic process, we repeated each experiment five times and we report the average performance. Also, as Labeled LDA produces a ranking of labels for each instance, we applied the Metalabeler approach [28] as a thresholding technique, in order to obtain the necessary bipartitions. The same model, a linear regression model, was used for both LLDA and HOMER-LLDA in the experiments.

Finally, all experiments were run on a machine with 50 Intel Xeon processors at 2.27GHz each and on 1Tb of RAM memory.

### 5.2 Data sets

For the first experiment, we used a small subset of the $BioASQ$ dataset [29]. The $BioASQ$ challenge deals with the semantic annotation of scientific articles from the biomedical domain. For each article, the abstract, the title, the journal and the year of publication are given, along with a list of MeSH tags, provided by the National Library of Medicine. For this experiment, we used the last 12,000 documents of the corpus, keeping the first 10,000 for training and the rest for testing. Stop-words and features with less than 5 occurrences were filtered out.

For the next three series of experiments, we employed $Bibtex$ [30], $Bookmarks$ [30], $EUR\text{-}Lex$ [31] and $Yelp$. The first three data sets have been extensively used in a number of papers, therefore we will not further describe them. For the $Yelp$ data set, we retrieved the data available from the Yelp Dataset Challenge website[5] and formulated a multi-label learning problem where the goal is to predict the attributes for each business by using text from the relevant reviews. More specifically, we obtained 1,569,265 reviews for a total of 61,185 businesses and after filtering out businesses with less than two reviews, we concatenated reviews for each of the remaining businesses resulting in a corpus of 56,797 instances. The label set of the problem consists of the set of attributes (only those with one value, Boolean, numerical or text) and the categories of each business (e.g. an instance's label set could consist of the following labels: "By Appointment Only","Price Range_2","Nail Salons","Accepts Credit Cards","Beauty & Spas"). Stop-words and words with less than 10 occurrences in the corpus were removed, yielding a set of 70,180 features in total.

For the fifth set of experiments, we used two large-scale data sets, $BioASQ$ and $DMOZ$ [32]. For $BioASQ$ we used a subset of the entire corpus (the last 250,000 documents) following the same preprocessing procedure as for the first experiment. In case of $DMOZ$ we did not perform any further processing, keeping 80% of the data for training and the rest for testing.

Table 2 shows the relevant statistics for all of the aforementioned datasets.

### 5.3 Evaluation measures

We employ two widely-used evaluation measures to assess performance: the micro-averaged and macro-averaged F1 measures (Micro-F and Macro-F, for short) [33]. These measures are a weighted function of precision and recall, and emphasize the need for a model to perform well in terms of both of these underlying measures. The Macro-F score is the average of the F1-scores that are achieved across all labels, and the Micro-F score is the average F1 score weighted by each label's frequency. Equations 9 and 10 provide the definitions of the two measures in terms of the true positives ($tp_l$), false positives ($fp_l$) and false negatives ($fn_l$) of each label $l$.

$$Micro-F1_{score} = \frac{2 \times \sum_{l=1}^{|L|} tp_l}{2 \times \sum_{l=1}^{|L|} tp_l + \sum_{l=1}^{|L|} fp_l + \sum_{l=1}^{|L|} fn_l} \quad (9)$$

$$Macro-F1_{score} = \frac{1}{|L|} \sum_{l=1}^{|L|} \frac{2 \times tp_l}{2 \times tp_l + fp_l + fn_l} \quad (10)$$

### 5.4 Effect on frequent and rare labels when increasing total number of nodes

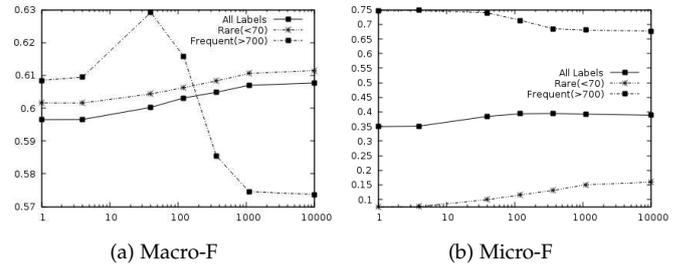

(a) Macro-F  (b) Micro-F

Fig. 3: Performance for rare and frequent labels of a subset of the $BioASQ$ dataset against the number of total nodes (in log scale) for seven different $HOMER$ models.

In this experiment, we want to study how well frequent and rare labels are learned from a HOMER model when we increase the total number of nodes in the hierarchy. We used a subset of the $BioASQ$ dataset, as this dataset has very few frequent labels and a great number of extremely rare ones and is therefore particularly suited for this empirical study. For the purpose of this experiment we considered as frequent, the labels having a frequency greater than 700 and as rare the labels with a frequency lower than 70. Figure 3 depicts the performance for rare and frequent labels for seven different $HOMER$ models in terms of Micro-F and Macro-F. The configuration for this models was as follows: BR SVMs were employed as the baseline MLC and for constructing the label hierarchy we used balanced $k$ means with $k = 3$ and $nmax = 3, 20, 100, 300, 1000, 10000, 20000$ for each of the models. This led to seven different models with 9841, 1093, 364 121, 40, 4 and 1 total nodes accordingly.

The results seem to validate our analysis in Section 4.4. Rare labels tend to benefit as the hierarchy becomes deeper and fewer labels per terminal node are observed. This is expected, as the sub-sampling process that HOMER follows is smoothing out the class imbalance problem for rare labels. On the contrary, frequent labels exhibit an inverse behavior; as the number of total nodes increases, the initial increase in performance is followed by a significant deterioration, for both measures. Again, this behavior is explained by considering the fact that frequent labels will have fewer and fewer negative examples as we create deeper hierarchies and, at some point, this will lead to an inverse imbalance effect, where the label will have many positive and very few negative examples.

---

5. http://www.yelp.com/dataset_challenge

TABLE 2: Data Sets Statistics.

| Data set | Documents | | | Labels | | | Word Types |
|---|---|---|---|---|---|---|---|
| | Training | Test | Average Length | Set | Cardinality | Average Frequency | |
| Bibtex | 4,880 | 2,515 | 68.46 | 159 | 2.38 | 73.05 | 1,479 |
| Bookmarks | 70,000 | 17,855 | 125.49 | 208 | 2.03 | 682.90 | 2,100 |
| EUR-Lex | 15,314 | 4,000 | 1274.19 | 3,826 | 5.29 | 21.17 | 26,575 |
| Yelp | 45,000 | 11,797 | 3531.23 | 814 | 9.76 | 539.75 | 70,180 |
| BioASQ(1st exp) | 10,000 | 2,000 | 211.52 | 13,283 | 13.12 | 107.89 | 19,145 |
| BioASQ | 200,000 | 50,000 | 221.68 | 24,094 | 13.53 | 112.36 | 92,293 |
| DMOZ | 322,465 | 72,288 | 358.34 | 27,689 | 1.03 | 11.97 | 108,230 |

The above experiments could serve as a generic guide to properly configure a HOMER model; when dealing with problems with many rare labels, we should aim at creating hierarchies with small $nmax$ values and therefore more total nodes. On the contrary, problems dominated by frequent labels, would lead us in choosing larger $nmax$ values.

As a final note for future extensions, we could consider the case of a model that would use a heuristic criterion to stop the partitioning of a given node, if the node has many frequent labels (or in the general case, if performance is expected to drop by further partitioning the node).

### 5.5 Empirical study on the number of clusters

In this set of experiments, we investigate how the number of clusters into which the labels of each node are partitioned affects the algorithm's performance. We select balanced $k$ means as the clustering algorithm (this is convenient, as this algorithm allows us to set explicitly the number of clusters), BR-SVM as the MLC and set the $nmax$ parameter accordingly for each of the used datasets. Specifically, we set $nmax$ to 20 for $Bibtex$, 10 for $Bookmarks$, 200 for $EUR$-$Lex$ and to 20 for $Yelp$. In Figure 4 and Figure 5 we report the results for different choices of the parameter, in terms of Macro-F and Micro-F respectively. The performance of the baseline method (BR-SVM) is also depicted to facilitate comparisons.

First, HOMER-BR has a steady advantage over BR across the different datasets and the various $k$ configurations. Only in one case out of the eight plots, for $Yelp$ in terms of Macro-F, we can observe BR being steadily better. Also, we can notice HOMER-BR getting worse than BR for $k = 10$, in one case for Macro-F and two for Micro-F, which suggests that for this configuration the constructed label hierarchy is of inferior quality.

Secondly, for both measures, we observe a similar tendency in three out of four datasets. As the number of clusters increases, performance has a declining trend, dropping even below the baseline for larger values. The fourth dataset, $Yelp$, has a different behavior being relatively steady in terms of Macro-F and improving Micro-F as the number of clusters increases. We note here that the number of clusters is essentially the primary factor of how the labels will be arranged in the hierarchy. For instance a large number of clusters will lead to a very "open" and shallow tree, while a smaller one will lead to deeper and more "closed" ones. Therefore, this is the main parameter that will affect the clustering's quality and subsequently performance and should be the first element to be considered for experimentation when seeking to construct an optimal clustering of the labels.

Moreover, even if these empirical results tend to favor a small number of clusters (two or three) as a safe default choice for configuring a HOMER model (the results in [5], Section 5.1 suggest as well a similar option) we advise against choosing a default option for this parameter as it is crucial for the quality of the resulting label hierarchy.

A number of clustering algorithms, e.g. density-based algorithms such as DBSCAN [34] or OPTICS [35] do not allow explicit setting of the number of clusters. The relevant parameters of each algorithm however, control indirectly as well the number of clusters and eventually the structure of the tree and should therefore be chosen carefully for optimal results.

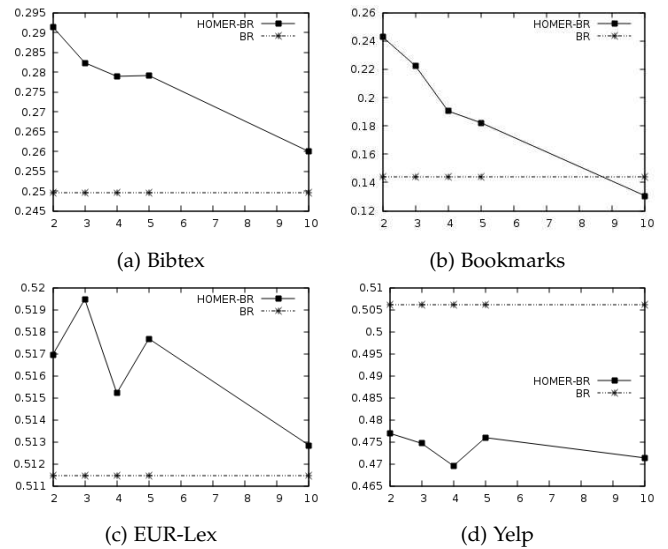

Fig. 4: $HOMER$-$BR$ results for five different choices of parameter $k$ for the four data sets, in terms of the Macro-F measure. The respective performance of $BR$ is also shown to visualize the improvement.

### 5.6 Empirical study of the $nmax$ parameter

In this experiment, we investigate the role of parameter $nmax$ in HOMER's performance. The above parameter controls the maximum allowed number of labels in a leaf node. This parameter essentially determines if a node will be further partitioned in a set of children nodes. In the initial algorithm presentation [5] all leaf nodes consisted of one label, the equivalent of setting $nmax = 1$. We ran HOMER on four data sets ($Bibtex$, $Bookmarks$, $Yelp$, $EUR$-$Lex$) for multiple $nmax$ values. We set the clustering algorithm to balanced $k$ means and the MLC to Binary Relevance with Linear SVMs. The parameter $k$ was fixed to 3 across all datasets. We present the results of this experiments in Figure 6 and 7. The performance of the baseline method (BR-SVM) is also depicted to facilitate comparisons.

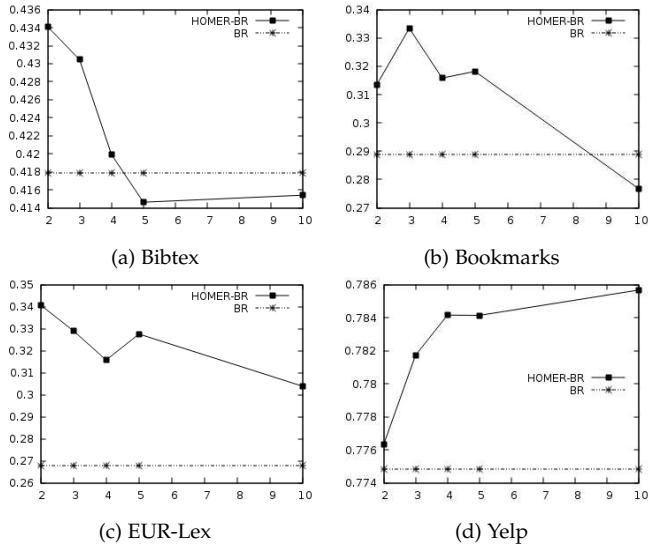

Fig. 5: HOMER-BR results for different choices of parameter $k$, for the four data sets, in terms of the Micro-F measure.

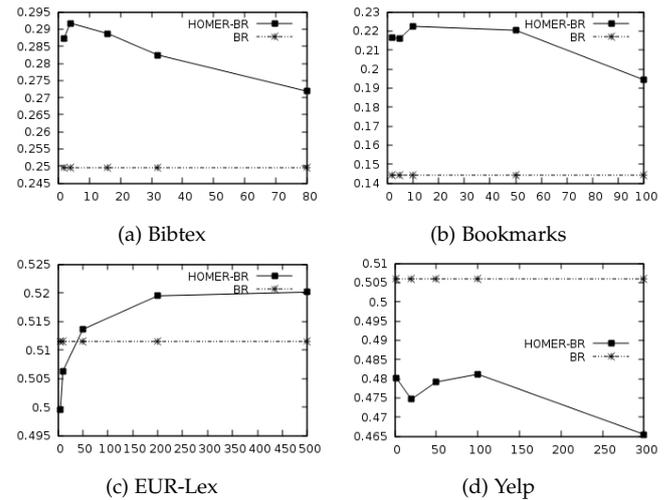

Fig. 6: HOMER-BR results for different choices of parameter $nmax$, for the four data sets, in terms of the Macro-F measure.

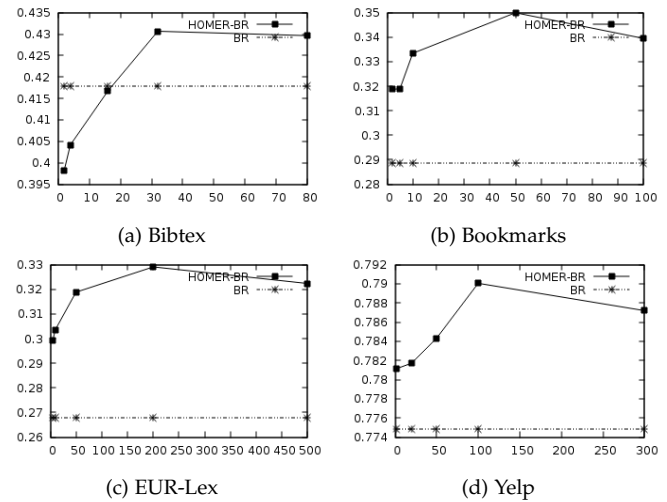

Fig. 7: HOMER-BR results for different choices of parameter $nmax$, for the four data sets, in terms of the Micro-F measure.

First, compared to the base MLC, HOMER-BR has once more the upper hand in the seven out of eight figures (apart from $Yelp$ in terms of Macro-F). Furthermore, for small values of $nmax$ in one case for Macro-F and in one case for Micro-F, HOMER's performance is worse than that of BR.

Secondly, there is a common trend in seven out of the eight figures (apart from the case of $EUR$-$Lex$ for Macro-F) with performance increasing initially with $nmax$, reaching a maximum value and then dropping again. As we have explained in Section 4.4, this is an expected behavior since, as $nmax \rightarrow 1$ the training sets of terminal nodes will be rather small, leading to a drop in performance. On the other hand, as $nmax \rightarrow |L|$, performance will tend to approach that of the base MLC with the hierarchy degenerating to a single cluster for $nmax = |L|$. These results validate the observations made in Section 4.4 about the fact that we should expect an optimal value to exist for the $nmax$ parameter.

### 5.7 Configuration paradigms; Different Clustering Algorithms and Classifiers

The goal of this series of experiments is to illustrate the ability of the described HOMER algorithm to accommodate various clustering algorithms and multi-label classifiers. Here, we describe six different instantiations of HOMER to serve as such example configurations, by employing three different clustering algorithms (balanced $k$ means described in Section 3, FastOPTICS [36] and SLINK [37]) and two different multi-label classifiers (BR-SVMs and Labeled LDA). For this experiment, we used four data sets, $Bibtex$, $Bookmarks$, $EUR$-$Lex$ and $Yelp$. For Labeled LDA, we employed the algorithm's extension described in Section 4.2. Also, the $k$ parameter described earlier is not valid in case of SLINK and FastOPTICS, as these two algorithms take different parameters. Specifically, for SLINK we kept default parameters and for FastOPTICS we set $\epsilon = 0.001$ and $minPts = nmax$. For these two clustering algorithms we followed again the approach that if a resulting cluster would have more than $nmax$ labels, then the algorithm would be run again on that cluster (in this case the $\epsilon$ parameter of FastOptics was doubled in order to allow for smaller clusters). We note that for the algorithm-specific parameters as well as for the $nmax$ parameter we did not perform an exhaustive search for the optimal parameters.

Table 3 shows the results for this round of experiments,

the run time for each experiment and, specifically for the HOMER models, the average training set size and number of total nodes.

First, let us examine the respective results for BR and HOMER-BR. All HOMER models demonstrate, overall, an improved performance over BR, with the exception of the Macro-F measure in $Yelp$ dataset. Balanced $k$ means is somewhat more consistent in outperforming the base MLC, while results for FastOPTICS and SLINK seem more mixed. When comparing LLDA and HOMER-LLDA, we observe more mixed results with LLDA having the upper hand in terms of Macro-F in three out of four datasets. In terms of Micro-F however, HOMER with balanced $k$ means is outperforming LLDA in all cases, while results for the other two clustering configurations appear again more diversified.

Even if comparisons among HOMER models should be taken with a grain of salt, given that we did not choose optimal parameters for each of the clustering algorithms and that each algorithm creates a hierarchy with a different structure and a different total number of nodes, we can remark that balanced $k$ means is performing consistently better than FastOPTICS and SLINK. Apart from the aforementioned factors, a possible reason for this behavior could be the fact that balanced $k$ means produces a balanced hierarchy. Similar results from the experiments in [5] may suggest that imposing such an explicit constraint of even distribution of labels among the nodes of the hierarchy, can perhaps affect significantly performance.

In Table 3 we additionally provide the running times for each model. To facilitate our analysis of the results, we also provide the average training corpus size for non-leaf nodes (depicted as $\overline{|D_{NL}|}$) and leaf nodes (depicted as $\overline{|D_L|}$) and the total number of nodes for each of the algorithms. If we examine the training times for the HOMER models that employ balanced $k$ means, we can observe that results are aligned with Equation 6 and the relevant conclusions of Section 4.3, with roughly similar times to BR-SVMs (apart from the $Yelp$ dataset for HOMER-BR). We note that HOMER includes also a filtering step of training instances from parent to child node and this step was not optimized in our code. Therefore, the differences we observe may also be due to that step. In case of the HOMER models using the rest of the clustering algorithms, conclusions from Section 4.3 do not apply as the latter perform unbalanced clustering. Nevertheless, we can observe in general a similar behavior with approximately equivalent training times of the HOMER models to the respective MLCs.

In case of the $Yelp$ dataset the training times are significantly longer than those observed for the given MLC. A possible reason for this could be the fact that the average label frequency is higher compared to the other datasets and subsequently this leads in bigger $|D_n|$ compared to $|D|$ and therefore longer training times. The number of total nodes also seems to play a role for unbalanced algorithms; SLINK tends to produce far bigger hierarchies than the other algorithms paying the price in terms of training duration.

Concerning prediction times, as we explained in Section 4.3, the computational complexity of a HOMER model in the case of a balanced clustering algorithm can vary from logarithmic in the best case, to linear in the worst case (if all paths of the label hierarchy are followed). Therefore, we see generally significantly shorter times for the HOMER models compared to the base MLC, a tendency not being limited to those models that employ balanced $k$ means. In two cases however (for $Bibtex$ and $EUR-Lex$) , SLINK has significantly longer times than the base MLC. A possible explanation could relate with the quality of the clustering; it seems that in this case the instances to be predicted are forwarded in large portions of the tree, causing the longer prediction times.

Overall, the results suggest that it is totally valid to employ any given clustering algorithm to construct the label hierarchy in the HOMER framework. In some cases (for instance in $Bibtex$ and $Yelp$ in terms of Micro-F), alternatives to balanced $k$ means can perform even better, therefore one should not rely on a default HOMER setup for a specific multi-label task. Another remark we could make [6], is that HOMER's performance, both in terms of running time and quality of prediction, seems to be largely dependent on the quality of the label clustering. In other words, the *Achilles' heel* of the algorithm described in this paper seems to be the choice of the given clustering algorithm's parameters.

## 5.8 Large-scale tasks

In the last round of experiments, we study two large-scale multi-label classification tasks, $BioASQ$ and $DMOZ$. We choose balanced $k$ means as a clustering algorithm to create the label hierarchy and BR-Linear SVMs as the multi-label classifier. Apart from performance, in this experiment we are also interested on training and prediction duration. Table 4 shows the relevant results. We also show the respective running times for the algorithms and $\overline{|D_n|}$ compared to $|D|$, as a means to illustrate the improvement in training complexity. The total number of nodes per model is also depicted.

These two multi-label tasks provide a characteristic example case where HOMER can bring a significant improvement both in performance and running times. First, in terms of both measures and for both tasks we observe a statistically significant gain in performance. Second, in terms of training times we also notice a significant improvement. Especially for $DMOZ$, training with HOMER-BR is almost sixteen times faster than with BR. This difference is partly due to the nature of the dataset; as $DMOZ$ has $\overline{|L_d|} \simeq 1$, the resulting $\overline{|D_n|}$ will be a lot smaller than $|D|$ allowing for faster training. Prediction is also conducted much faster, at half time for $BioASQ$ and one third of the time for $DMOZ$.

These results may provide a hint on when HOMER is more appropriate to be employed on a multi-label task. Applications with large $|L|$ and $|D|$ appear to be more suitable, rendering the application of a given MLC more beneficial at the same time improving the relevant running times.

## 6 CONCLUSION

In this work we have presented the HOMER framework, an approach that can wrap any given multi-label classifier, with the aim to improve on performance and running time. The

---
6. initial experimentation on the datasets used throughout the paper validated these observations.

IEEE TRANSACTIONS ON KNOWLEDGE AND DATA ENGINEERING,, VOL. 0, NO. 0, 210011

TABLE 3: Results on $Bibtex$ and $Bookmarks$

| Data set | Classifier | Clusterer | Performance Micro-F | Macro-F | Duration Training | Test | $\overline{|D_{NL}|} + \overline{|D_L|}$ | nodes |
|---|---|---|---|---|---|---|---|---|
| Bibtex | | | | | | | | |
| | BR | | 0.41783 | 0.24953 | 0.5 | 0.1 | 4,880 | |
| | H-BR | $k$ means (3, 40) | **0.43068** | **0.28247**△ | 0.6(0) | 0 | 3,206.2+1,140.8 | 13 |
| | H-BR | FastOPTICS (40) | 0.42346 | 0.26977△ | 1.1(0.8) | 0 | 4,880+2,044.2 | 5 |
| | H-BR | SLINK (40) | 0.42202 | 0.27034△ | 0.5(0.1) | 0 | 4,880+212.6 | 43 |
| | LLDA | | 0.37517 | 0.24802 | 0.4 | 0.6 | 4,880 | |
| | H-LLDA | $k$ means (3, 40) | 0.37668 | 0.239723 | 0.4(0) | 0.3 | 3,206.2+1,140.8 | 13 |
| | H-LLDA | FastOPTICS (40) | **0.39018** | **0.24820** | 1.1(0.8) | 0.3 | 4,880+2,044.2 | 5 |
| | H-LLDA | SLINK (40) | 0.36628 | 0.20726△ | 0.5(0.1) | 3.0 | 4,880+212.6 | 43 |
| Bookmarks | | | | | | | | |
| | BR | | 0.28875 | 0.14408 | 16.1. | 1.2 | 70,000 | |
| | H-BR | $k$ means (3, 10) | **0.33339**△ | **0.22248**△ | 18.4(4.3) | 0.1 | 23,075.3+5,048.9 | 40 |
| | H-BR | FastOPTICS (10) | 0.31856△ | 0.16697△ | 45.6(26.2) | 0.8 | 51,057.5+9,749 | 14 |
| | H-BR | SLINK (10) | 0.30639△ | 0.16530△ | 20.5(4.8) | 0.3 | 33,289.3+2,304.5 | 54 |
| | LLDA | | 0.20144 | **0.11251** | 17.5 | 20.2 | 70,000 | |
| | H-LLDA | $k$ means (2, 55) | **0.20460** | 0.09594△ | 19.2(3.1) | 8.2 | 53,804.3+27,779.5 | 7 |
| | H-LLDA | FastOPTICS (55) | 0.16744△ | 0.06793△ | 27.5(17.2) | 8.6 | 70,000+18,345 | 7 |
| | H-LLDA | SLINK (55) | 0.19774 | 0.07784△ | 14.0(3.5) | 5.6 | 70,000+4,868.4 | 24 |
| EUR-Lex | | | | | | | | |
| | BR | | 0.26793 | 0.51146 | 36.8 | 1.1 | 15,314 | |
| | H-BR | $k$ means(3, 200) | **0.32908**△ | **0.51947** | 23.4(5.3) | 0.9 | 9,062.5+2,764.7 | 40 |
| | H-BR | FastOPTICS(200) | 0.29712△ | 0.51134 | 82.4(49.3) | 2.1 | 15,314+3,199.5 | 20 |
| | H-BR | SLINK(200) | 0.27052 | 0.50964 | 144.1(49.1) | 4.0 | 15,314+63.5 | 966 |
| | LLDA | | 0.10917 | **0.44309** | 61.8 | 169.3 | 15,314 | |
| | H-LLDA | $k$ means(3, 500) | **0.11657** | 0.43355 | 67.9(4.4) | 66,4 | 13,682+7,009.4 | 13 |
| | H-LLDA | FastOPTICS(500) | 0.08705△ | 0.38074△ | 80.6(41,1) | 50.4 | 15,314+3673.2 | 14 |
| | H-LLDA | SLINK(500) | 0.08344△ | 0.43538 | 100.5(46.4) | 42.0 | 15,314+79.4 | 976 |
| Yelp | | | | | | | | |
| | BR | | 0.77480 | **0.50605** | 60.4 | 7.3 | 45,000 | |
| | H-BR | $k$ means (3, 20) | 0.78371△ | 0.47463△ | 228.4(28.4) | 6.1 | 19,155.3+5,201.4 | 121 |
| | H-BR | FastOPTICS (20) | 0.78690△ | 0.46959△ | 188.3(77.2) | 8.0 | 32,361+10,985.5 | 17 |
| | H-BR | SLINK (20) | **0.78693**△ | 0.46878△ | 119.7(10.3) | 7.1 | 15,900+913.9 | 180 |
| | LLDA | | 0.59825 | **0.32851** | 135.4 | 169.4 | 45,000 | |
| | H-LLDA | $k$ means (3, 100) | **0.61840**△ | 0.32642 | 131.3(11.1) | 38.0 | 41,946.2+26,304.3 | 13 |
| | H-LLDA | FastOPTICS (40) | 0.29194△ | 0.29757△ | 327.3(75.1) | 81.1 | 45,000+9,623.1 | 18 |
| | H-LLDA | SLINK (100) | 0.28436△ | 0.27729△ | 381.7(14.5) | 42.3 | 45,000+963.3 | 172 |

[1]HOMER models are denoted with H-MLC. For each different clustering technique, we show the exact parameterization in parentheses, the first number denoting $k$ and the second $nmax$. The △ symbol represents a statistically significant difference between the base MLC and the respective HOMER model at $p = 0.05$ (we use the symbol either if the HOMER model is significantly better or significantly worse than the MLC of choice). In the 'Duration' column, figures are given in minutes, a 0 noting a duration of less than 6 seconds. In the 'Training' column, the first number concerns the total training time while the number in parentheses the clustering time.

TABLE 4: Results on $BioASQ$ and $DMOZ$

| Data set | Classifier | Performance Micro-F | Macro-F | Duration Training | Test | $\overline{|D_n|}$ | nodes |
|---|---|---|---|---|---|---|---|
| BioASQ | | | | | | | |
| | BR | 0.54282 | 0.40644 | 784 | 33 | 200,000 | |
| | HOMER-BR ($k = 3$, $nmax = 800$) | **0.55061**△ | **0.41646**△ | 452.4(59) | 15 | 49,906.3+15,432.3 | 121 |
| DMOZ | | | | | | | |
| | BR | 0.20512 | 0.24413 | 3,688 | 24 | 322,465 | |
| | HOMER-BR($k = 3$, $nmax = 500$) | **0.24146**△ | **0.26288**△ | 232(66) | 8 | 32707.2+4078.1 | 121 |

algorithm breaks down the global multi-label task to several smaller subtasks, by first employing recursively a clustering algorithm on the label set, creating a label hierarchy. Training and prediction is subsequently carried out locally at each node of the hierarchy. The algorithm has a linear training complexity and a logarithmic testing complexity, irrespective of the employed MLC.

The empirical results from the experiments carried out in this paper, demonstrate that HOMER can significantly improve performance when applied on a given MLC method. Special care should be given however, to adjust optimally the clustering algorithm's parameters, as this is the part that affects most the algorithm's behavior.

Specifically for the last part of the experiments, the positive results may indicate that HOMER is especially apt in addressing large-scale multi-label tasks.

Finally, as a future extension of this work, we would like to consider possible extensions of HOMER, to address Extreme Classification tasks, multi-label tasks with hundreds of thousands of labels or more.

[1] L. Schietgat, C. Vens, J. Struyf, H. Blockeel, D. Kocev, and S. Džeroski, "Predicting gene function using hierarchical multi-label decision tree ensembles," *BMC bioinformatics*, vol. 11, no. 1, p. 2, 2010.

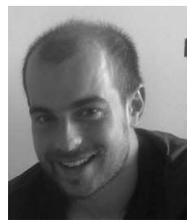

**Yannis Papanikolaou** received in 2008 his diploma (5-year degree) from the school of Electrical and Computer Engineering, National Technical University of Athens. He then worked for 4 years as an informatics teacher in high-school. Since 2014, he is pursuing a PhD in text mining, in the School of Informatics of Aristotle University of Thessaloniki under the supervision of Grigorios Tsoumakas. His scientific interests include multilabel learning, text mining and (un)supervised learning.

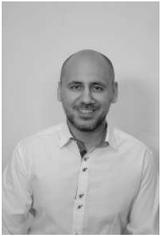

**Ioannis Katakis** is a senior researcher at the National and Kapodistrian University of Athens working on EU-funded research projects as a Quality Assurance Coordinator. Born and raised in Thessaloniki, Greece, studied Computer Science and holds a PhD in Machine Learning. After his post-graduate studies, he served various universities in Greece and Cyprus as a lecturer and researcher (Aristotle University, University of Cyprus, Cyprus University of Technology, Open University of Cyprus, Hellenic Open University, Athens University of Economics and Business). There, he taught Machine Learning, Data Mining and Artificial Intelligence, while he supervised a large number of MSc and BSc theses. His research interests include Mining Social and Web Data, Knowledge Discovery from Text and Urban Data Streams, Multi-label Learning, Adaptive and Recommender Systems. He has published papers in International Conferences and Scientific Journals related to his area of expertise (e.g. ECML/PKDD, IEEE TKDE, ECAI), organized three workshops (at ICML, ECML/PKDD, EDBT/ICDT), edited three special issues (DAMI, InfSys) and is an Editor at the journal Information Systems. He regularly serves the programme committee of conferences (ECML/PKDD, WSDM, DEBS, IJCAI) and evaluates articles in numerous journals (TPAMI, DMKD, TKDE, TKDD, JMLR, TWEB, ML).

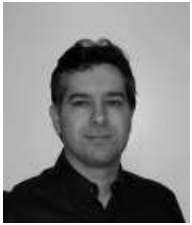

**Grigorios Tsoumakas** is an Assistant Professor of Machine Learning and Knowledge Discovery at the Department of Informatics of the Aristotle University of Thessaloniki (AUTH) in Greece. He received a degree in computer science from AUTH in 1999, an MSc in artificial intelligence from the University of Edinburgh, United Kingdom, in 2000 and a PhD in computer science from AUTH in 2005. His research expertise focuses on supervised learning techniques (ensemble methods, multi-target prediction) and text mining (semantic indexing, sentiment analysis, topic modeling). He has published more than 80 research papers and according to Google Scholar he has more than five thousand citations and an h-index of 30. Dr. Tsoumakas is a member of the Data Mining and Big Data Analytics Technical Committee of the Computational Intelligence Society of the IEEE and a member of the editorial board of the Data Mining and Knowledge Discovery journal.